\documentclass{article}
\usepackage{spconf,amsmath,graphicx,hyperref}
\usepackage{pdfpages}
\usepackage{afterpage}
\usepackage{algorithm}
\usepackage{algpseudocode}
\usepackage{algorithmicx}
\usepackage{subcaption}
\usepackage{float}
\usepackage{stfloats}

\newcommand{\argmin}{\mathop{\mathrm{argmin}}}
\title{DIR-TIR: Dialog-Iterative Refinement for Text-to-Image Retrieval}
\name{
  Zongwei Zhen\quad 
  Biqing Zeng$^{\star}$\thanks{Corresponding author: zengbiqing137@163.com}
}
\address{
    South China Normal University \\
    School of Artificial Intelligence \\
  Foshan, China \\
  }
\begin{document}
%
\maketitle
\begin{abstract}
    This paper addresses the task of interactive, conversational text-to-image retrieval.
    Our DIR-TIR framework progressively refines the target image search through two specialized modules: the Dialog Refiner Module and the Image Refiner Module.
    The Dialog Refiner actively queries users to extract essential information and generate increasingly precise descriptions of the target image.
    Complementarily, the Image Refiner identifies perceptual gaps between generated images and user intentions, strategically reducing the visual-semantic discrepancy. By leveraging multi-turn dialogues, DIR-TIR provides superior controllability and fault tolerance compared to conventional single-query methods, significantly improving target image hit accuracy.
    Comprehensive experiments across diverse image datasets demonstrate our dialogue-based approach substantially outperforms initial-description-only baselines, while the synergistic module integration achieves both higher retrieval precision and enhanced interactive experience.
\end{abstract}
\begin{keywords}
    Text-to-Image Retrieval, Iterative Refinement, Interactive Dialog Systems, Conversational Image Search
\end{keywords}
\section{Introduction}
\label{sec:intro}

Image retrieval is a critical computer vision task, particularly in today's era of exponentially growing visual data repositories.\cite{ma2025multi}
As users grapple with massive image collections - many individuals routinely store thousands or even tens of thousands of photographs on personal mobile devices. The challenge of efficiently locating target images has reached unprecedented importance.
While considerable progress has been made through the development of vision-language multimodal models \cite{li2023blip,openai2024gpt4o,yin2024survey}, existing methods exhibit fundamental limitations in text-to-image retrieval:
such as CLIP \cite{radford2021learning} and ALIGN \cite{jia2021scaling} leverage contrastive language-image pretraining to achieve impressive zero-shot retrieval on benchmark datasets.
However, their reliance on unidirectional query processing renders them vulnerable to the "semantic bottleneck" effect: when initial descriptions are not very precise (e.g., "a dog is playing" vs. "golden retriever catching red frisbee in park"), the accuracy of retrieval will significantly decrease.
This necessitates the user to provide a comprehensive and detailed description of the target image.

Recently, Levy et al.\cite{levy2023chatting} introduced a chat-based image retrieval system that leverages large language models (LLMs)\cite{zhao2023survey,li2024survey,shao2024survey} to pose clarifying questions to users while fine-tuning the model's text encoder to accept conversational inputs.
Through this multi-round dialogue paradigm, retrieval efficiency and performance are enhanced even when users provide only brief initial image descriptions.
However, this conversational retrieval framework faces critical limitations-including its resource-intensive fine-tuning requirement to enable dialogue-style text encoding and poor scalability prospects.
Moreover, when utilizing the LLM to generate clarifying questions during the retrieval process, the proposed queries rely solely on the initial image description and dialogue history.
This constrained context often results in stochastically generated questions with significant randomness-many may be irrelevant to the target image.

To address these issues, this paper proposes DIR-TIR, a plug-and-play text-to-image retrieval framework with two core modules: the Dialog Refiner Module and the Image Refiner Module.
The Dialog Refiner Module iteratively refines image descriptions by generating targeted questions to make descriptions more accurate representations of the target image.
This process involves two key steps:
First, it leverages information from candidate image sets as contextual reference for generating discriminative questions via LLMs.
After receiving user responses, it then utilizes the LLM's reasoning capability to synthesize new image descriptions by integrating dialogue history with the original description.
Meanwhile, the Image Refiner Module uses a text-to-image generation model to create generated images.
Users then identify discrepancies between these generated images and the actual target image.
Using this visual feedback along with the original prompts, the module applies LLM reasoning to produce optimized generation prompts with improved accuracy.

\begin{figure*}[!t]

    \begin{minipage}[b]{1.0\linewidth}
        \centering
        \includegraphics[width=0.9\textwidth, keepaspectratio]{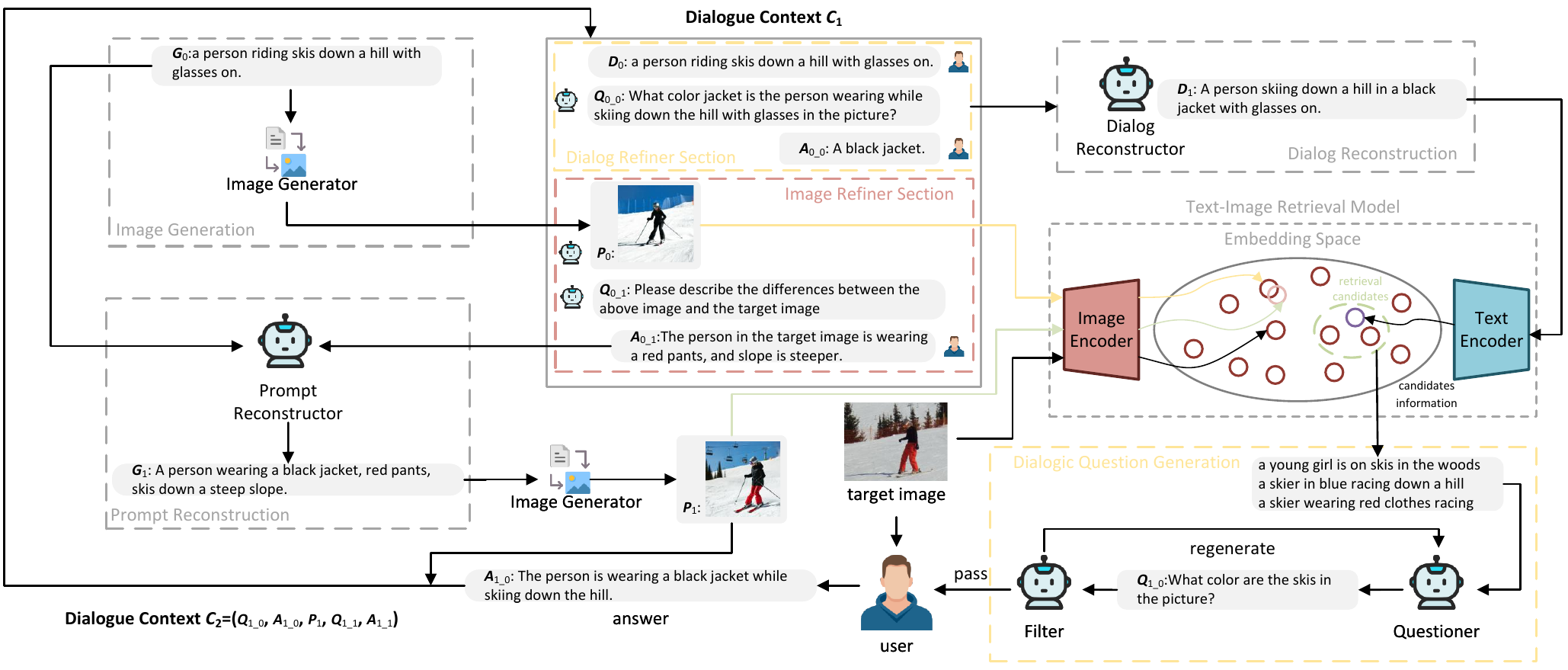}
    \end{minipage}
    \caption{The main framework of the DIR-TIR}
    \label{figure1}
\end{figure*}

Fig. \ref{figure1} presents the overall architecture of DIR-TIR, illustrating how both modules operate synergistically to achieve precise image retrieval through human-AI interaction.

\section{Method}
\label{sec:Method}

\subsection{Conversational text-to-image retrieval}

In conversational text-to-image retrieval, the user first provides an initial description $D_0$ of the target image they wish to find, initiating a multi-turn dialogue.
Denoting the current turn as $k$, in the $k$-th dialogue turn: The Dialog Refiner Module first poses a question $Q_{k\textunderscore0}$ to refine the image description.
The user then provides an answer $A_{k\textunderscore0}$ based on the content of the target image.
Next, the Image Refiner Module uses the prompt $G_k$ to generate an image $P_k$ through a text-to-image generation model.
Subsequently, the system asks $Q_{k\textunderscore1}$ to request the user to describe differences between $P_k$ and the target image, obtaining the user response $A_{k\textunderscore1}$.
This creates the dialogue record $C_k = (Q_{k\textunderscore0}, A_{k\textunderscore0}, P_k, Q_{k\textunderscore1}, A_{k\textunderscore1})$.
This paper constrains the maximum number of dialogue rounds to 10.

\subsection{The Dialog Refiner Module}

The Dialog Refiner Module implements an iterative description refinement mechanism designed to progressively align textual descriptions with the semantic content of target images.
The process begins with the user providing an initial description of the target image.
This description is then embedded into a high-dimensional semantic space using a retrieval model.
By calculating cosine similarity, we obtain a set of images that partially match the description, representing images sharing semantic similarities with the target to varying degrees.
Then, through clustering algorithms, these images are categorized into several major groups.
Information entropy metrics are then applied to identify the most representative image from each cluster.
Algorithm \ref{algorithm1} demonstrates the process of obtaining the contextual reference.

To further ensure that the generated questions effectively elicit new information, we incorporate an additional filtering step from recent work\cite{zheng2023ddcot}.
Using LLM judgments, we verify whether a proposed question can be answered with existing knowledge.
For each question, if the LLM agent cannot determine the answer based on the corresponding description and dialogue context, we prompt it to respond with 'Uncertain'.
Else if the LLM determines that the question cannot be sufficiently addressed with available information, it is deemed valid for extracting additional insights about the target image.

To address the proposed questions while circumventing the labor-intensive and costly process of human-machine dialogue collection, we employ a multimodal model to simulate user responses.
Given an input question and the target image, this model generates a ground-truth response based on the target image.
Subsequently, leveraging the reasoning capabilities of a large language model (LLM), we refine the original image description by incorporating the most recent question-answer pair.
This iterative process yields progressively more target-faithful image descriptions.
Finally, using the text encoder of our retrieval model, these enhanced descriptions are converted into high-dimensional semantic vectors, enabling cosine similarity computations against embeddings of all images in the entire dataset.

\begin{algorithm}[h]
    \caption{obtain contextual reference}
    \label{algorithm1}
    \begin{algorithmic}[1]
        \Require
        $D$: User description \\
        $r$: Current round number \\
        $N$: related\_size \\
        $\phi_t$: Text encoder \\
        $\phi_v$: Image encoder \\
        $K$: Number of clusters

        \Ensure $\mathcal{C}_{\text{rep}}$: Selected captions

        \State $v_d \gets \text{normalize}(\phi_t(D))$ \Comment{Embed description}
        \State $S \gets v_d \cdot V_{\text{img}}^\top$ \Comment{Compute similarities}
        \State $\text{topk} \gets \text{argsort}(S)[-N:]$ \Comment{Top-N indices}

        \State $\mathcal{E} \gets \text{zeros}(N)$ \Comment{Initialize entropies}
        \For{$i = 0$ to $N-1$}
        \State $v_c \gets \text{normalize}(\phi_t(\text{captions}[\text{topk}[i]]))$
        \State $P \gets \text{softmax}(v_c \cdot V_{\text{topk}}^\top)$
        \State $\mathcal{E}[i] \gets -\sum P \log P$ \Comment{Compute entropy}
        \EndFor

        \State $\text{clusters} \gets \text{KMeans}(V_{\text{img}}[\text{topk}], k=K)$
        \State $\mathcal{C}_{\text{rep}} \gets \emptyset$
        \For{$c = 0$ to $K-1$}
        \State $\text{indices} \gets \{j \mid \text{clusters}[j] = c\}$
        \If{$\text{indices} \neq \emptyset$}
        \State $m \gets \argmin_{j \in \text{indices}} \mathcal{E}[j]$ \Comment{Min entropy}
        \State $\mathcal{C}_{\text{rep}} \gets \mathcal{C}_{\text{rep}} \cup \{\text{captions}[\text{topk}[m]]\}$
        \EndIf
        \EndFor

        \State \Return $\mathcal{C}_{\text{rep}}$
    \end{algorithmic}
\end{algorithm}

\subsection{The Image Refiner Module}
The Image Refiner Module optimizes generative image prompts through iterative user feedback on visual discrepancies.
It begins with an initial prompt generating a original generated image.
Users then identify differences between this generated image and the target image.
Building upon the aforementioned approach, we similarly employ a multimodal model to simulate the user's role.
When provided with both the generated image and the target image, this model identifies differential features between the two images through guided prompting.
Both the original prompt and disparities are fed into the LLM, which synthesizes an improved prompt using its reasoning capabilities.
This refinement progress repeats so that the images achieve high visual similarity to the target.
Then, we use a retrieval model's image encoder to embed the generated image into a high-dimensional semantic space, enabling target image retrieval through cosine similarity computations.

\subsection{Integration of the results from the two modules}
The Dialog Refiner Module and the Image Refiner Module each generate independent ranked lists of all images in the gallery set based on similarity scores.
Higher-ranked images indicate higher confidence in being the target image.
For evaluation, we employ Hits@10 and Recall@10 as metrics:
Hits@10 indicates whether the target image appears in the top-10 retrieval results at any point up to and including the current dialogue round, measuring cumulative retrieval success.
Recall@10 assesses whether the target image ranks within the top-10 during the current round specifically, capturing immediate retrieval performance.
To form the final candidate image set, we implement a hybrid selection scheme that adaptively combines top-ranked images from both modules at varying ratios during each dialog round.
We define $dial\_num$ as the number of candidate images selected from the Dialog Refiner Module and $image\_num$ as those selected from the Image Refiner Module.
Their relationship is expressed by Equation \ref{eq:total_candidates}:
\begin{equation}
    dial\_num + image\_num = 10
    \label{eq:total_candidates}
\end{equation}
These complementary selections merge into a final set of 10 non-repetitive candidate images used for calculating both Hits@10 and Recall@10 scores.

\section{EXPERIENCE}
\label{sec:Experience}

\subsection{Experimental Settings}

We conduct text-to-image retrieval tasks on three benchmark datasets: VisDial\cite{szegedy2015proceedings}, COCO\cite{lin2014microsoft}, and Flickr30k\cite{young2014image}.
BLIP\cite{li2022blip} serves as the baseline retrieval model, while CLIP is additionally employed to further demonstrate that conversational refinement enables superior target image retrieval.
In the Dialog Refiner Module, we utilize DeepSeek-R1-Distill-Qwen-14B\cite{guo2025deepseek} for question generation, description refinement, and non-essential question filtering.
Qwen2.5-VL-7B\cite{Qwen2VL} functions as the user simulator for providing target-grounded responses.
In the Image Refiner Module, Stable-Diffusion-3.5\cite{esser2024scaling} serves as the text-to-image generation model.
Qwen2.5-VL-7B acts as the user simulator to identify visual discrepancies between generated images and the target reference.
DeepSeek-R1-Distill-Qwen-14B functions as the summarizer, generating refined image-generation prompts by integrating original prompts with discrepancy analyses.

We set K = 10 for k-means and keep it fixed, since larger K produces only redundant context and negligible gains in targeted question generation by the LLM.
\begin{figure*}[t!]
    \centering
    \begin{minipage}[t]{0.3\textwidth}
        \centering
        \includegraphics[width=\linewidth]{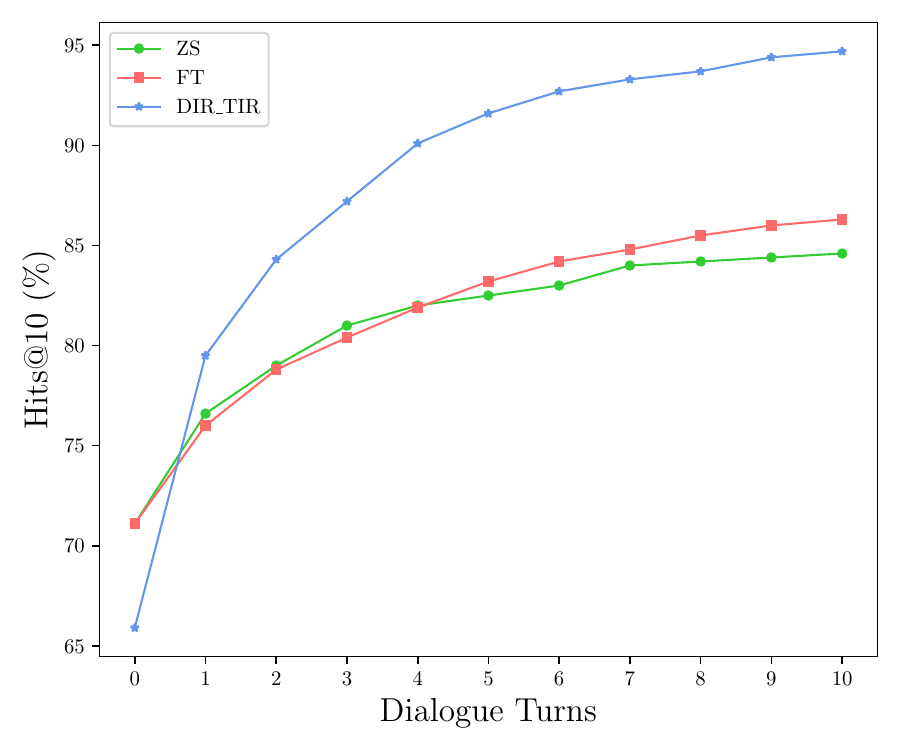}
        \subcaption{VisDial}
    \end{minipage}
    \hfill
    \begin{minipage}[t]{0.3\textwidth}
        \centering
        \includegraphics[width=\linewidth]{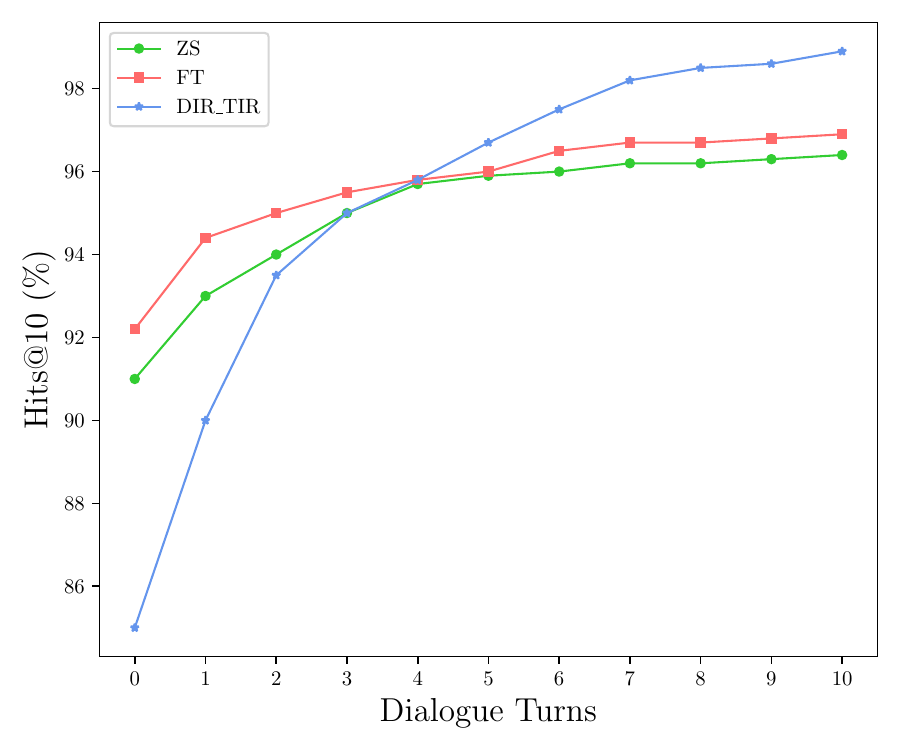}
        \subcaption{COCO}
    \end{minipage}
    \hfill
    \begin{minipage}[t]{0.3\textwidth}
        \centering
        \includegraphics[width=\linewidth]{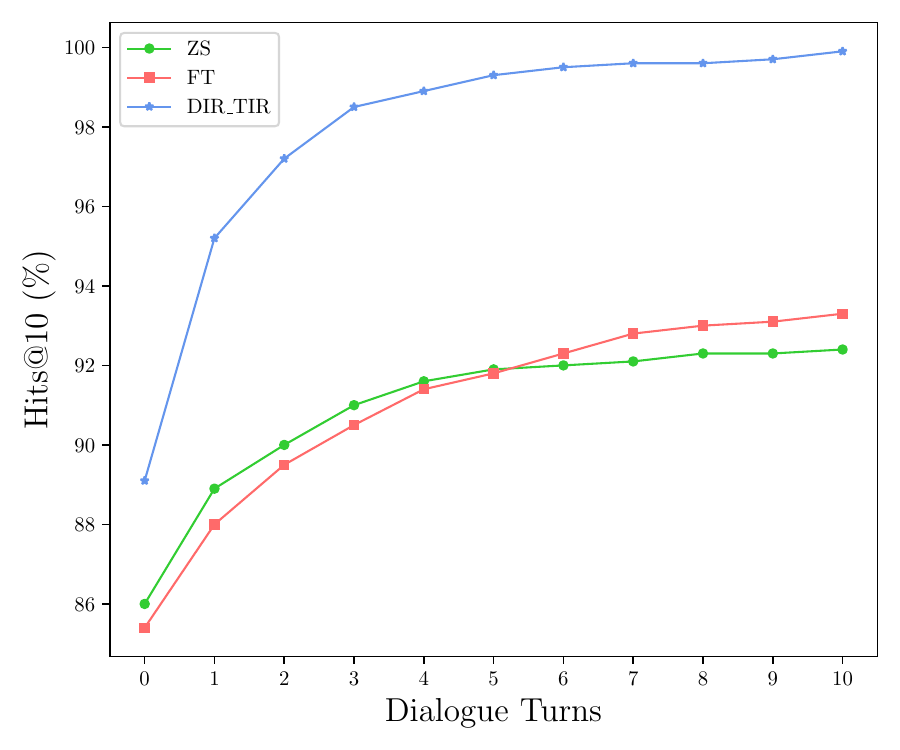}
        \subcaption{Flickr30k}
    \end{minipage}
    \caption{DIR-TIR comparison against ZS/FT methods on VisDial, COCO and Flickr30k.}
    \label{figure2}
\end{figure*}

\subsection{Results}

\begin{figure}[H] 
    \centering
    \begin{minipage}[t]{0.235\textwidth}
        \centering
        \includegraphics[width=\linewidth]{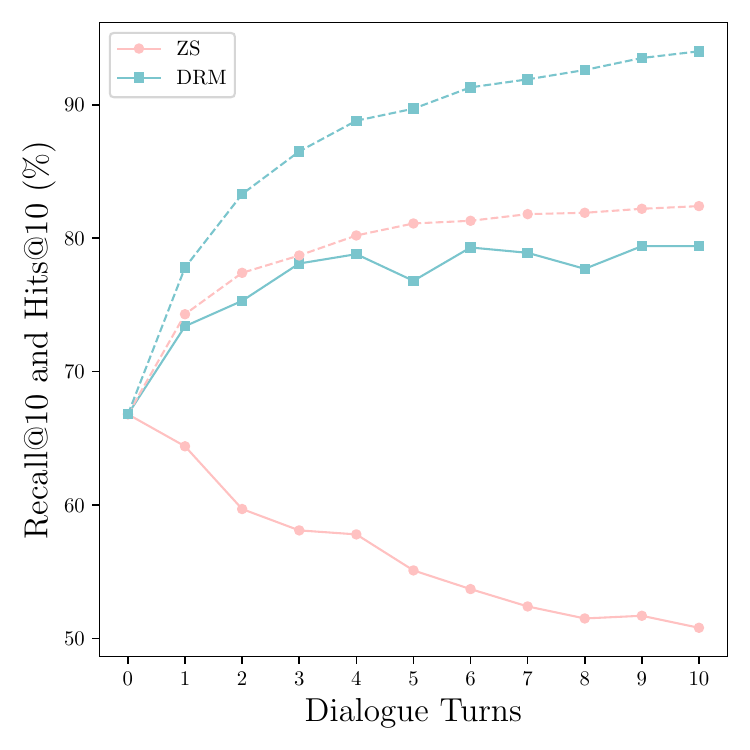}
        \subcaption{Blip as retrieval model}
    \end{minipage}
    \hfill
    \begin{minipage}[t]{0.235\textwidth}
        \centering
        \includegraphics[width=\linewidth]{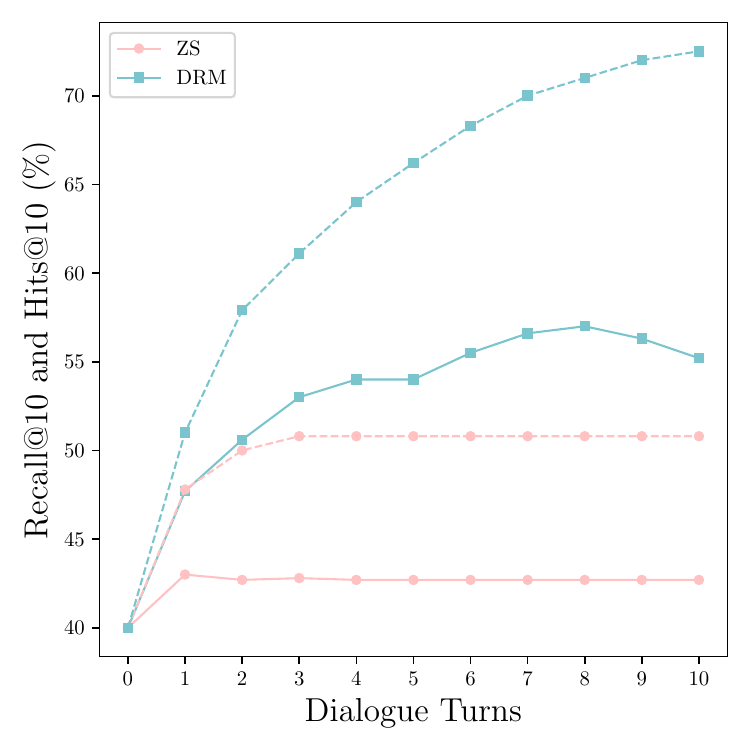}
        \subcaption{Clip as retrieval model}
    \end{minipage}
    \caption{Performance Comparison of Zero-Shot (ZS) vs Dialog Refiner Module (DRM) Implementations.
        Dashed lines show Hits@10 score, solid lines show Recall@10 score.}
    \label{figure3}
\end{figure}

To verify the intrinsic dialogue comprehension capability of retrieval models, we selectively isolate module contributions.
Following our hybrid framework design blending the Dialog Refiner and Image Refiner Modules, this configuration exclusively employs the Dialog Refiner Module outputs.
BLIP and CLIP models were evaluated across two distinct input conditions:using raw dialogue records for baseline and using Dialog Refiner processed contexts for enhanced retrieval.
Figure \ref{figure3} presents comparative Recall@10 and Hits@10 scores.

When using BLIP without dialog refinement, Recall@10 exhibits progressive degradation as dialogue turns increase.
CLIP baseline performance suffers from input truncation artifacts caused by inherent encoder length limitations.
Dramatic improvement emerges with the Dialog Refiner Module:both BLIP and CLIP achieve substantial gains in Hits@10 and Recall@10 metrics.

\begin{figure}[t]
    \centering
    \begin{minipage}{\linewidth}
        \includegraphics[width=0.9\columnwidth]{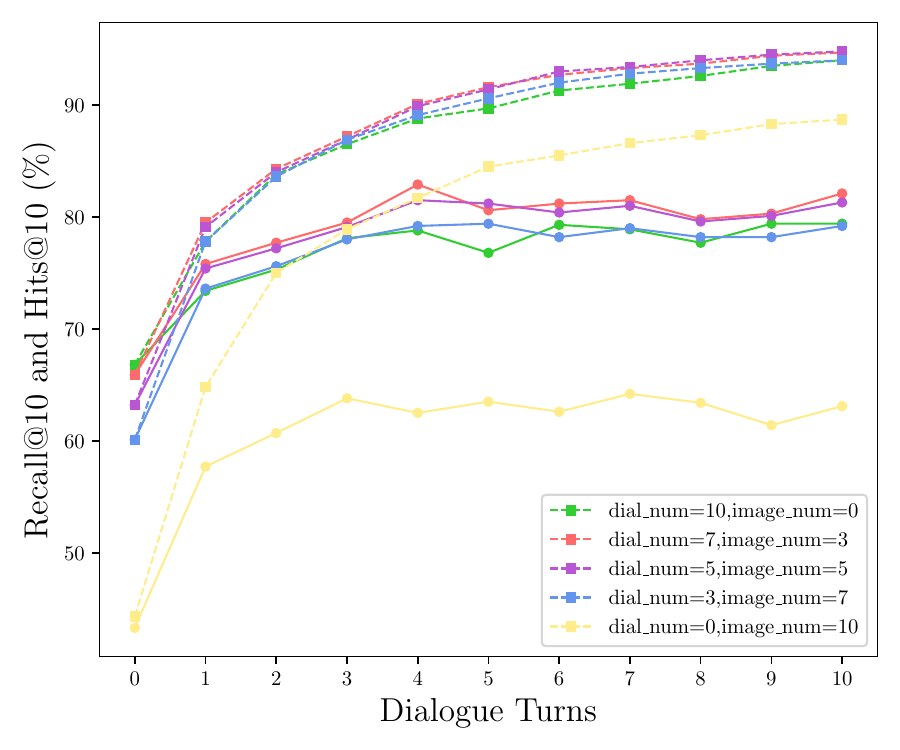}
        \captionof{figure}{Retrieval performance metrics (Recall@10 and Hits@10) achieved by hybrid compositions of the Dialog Refiner Module and Image Refiner Module at variable mixing ratios.
            Dashed lines show Hits@10 score, solid lines show Recall@10 score.}
        \label{figure4}
    \end{minipage}
\end{figure}

We systematically vary the proportion of results integrated from the Dialog Refiner and Image Refiner Modules to evaluate hybrid compositional effects.
Performance metrics (Hits@10 and Recall@10) across these ratio configurations are quantified in Figure \ref{figure4}, revealing the sensitivity of retrieval accuracy to module balance.

According to the experiment, the configuration $dial\_num=7$, $image\_num=3$ achieves the optimum balance between Hits@10 and Recall@10 metrics.
This parameter set is consequently adopted as the default composition ratio for final candidate image retrieval.

DIR-TIR outperforms multi-turn dialogue-based image retrieval baselines across VisDial, COCO, and Flickr30k (figure \ref{figure2}).
DIR-TIR demonstrates superior performance against comparable methods (ZS, FT\cite{levy2023chatting}) under identical BLIP retrieval backbones, ensuring fair evaluation.
We prioritize Hits@10 as the key metric, directly reflecting users' success rate in locating target images through conversational interaction.
According to the experiment, DIR-TIR outperforms baselines with steeper Hits@10 gains per dialogue turn, enhancing target image localization.

\section{Conclusion}

In this work, we present DIR-TIR, a plug-and-play framework that elevates text-to-image retrieval from single-shot matching to iterative, context-aware refinement.
By tightly coupling a Dialog Refiner Module with an Image Refiner Module, the system exploits both textual dialogue and pixel-level feedback without requiring costly fine-tuning of the underlying vision-language model.
Key technical insights are twofold: (i) candidate-image context serves as a discriminative prior that steers LLMs to ask highly-targeted questions.
(ii) visual discrepancy signals from user-annotated generated images provide an auxiliary gradient that refines the prompt in a semantic “closed-loop”, yielding continuous improvement across dialogue turns.




\bibliographystyle{IEEEbib}
\bibliography{strings,refs}

\end{document}